\documentclass{IOS-Book-Article}
\usepackage{mathptmx}
\usepackage{soul}\setuldepth{article}
\usepackage{graphicx}
\usepackage{longtable}
\usepackage{multirow}

\usepackage[table, svgnames, dvipsnames]{xcolor}
\usepackage{slashbox}
\usepackage{xspace}
\usepackage{makecell, cellspace, caption}

\begin{document}
\begin{frontmatter}              

\title{A Strategy for Implementing description Temporal Dynamic Algorithms in Dynamic Knowledge Graphs by SPIN}


\author[A]{\fnms{First} \snm{Alireza Shahbazi}
\thanks{Corresponding Author: Alireza Shahbazi, shahbazi@borhan-onto.ir}},
\author[B]{\fnms{Second} \snm{Seyyed Ahmad Mirsanei}},
\author[A]{\fnms{Third} \snm{Malikeh Haj Khan Mirzaye Sarraf}} and
\author[A]{\fnms{Fourth} \snm{Behrouz Minaei Bidgoli}}

\address[A]{School of Computer Engineering, Iran University of Science and Technology}
\address[B]{Department of Logic \& Philosophy, Tarbiat Modares University}

\begin{abstract}

\textbf{Purpose -}
Planning and reasoning about actions and processes, in addition to reasoning about propositions, are important issues in recent logical and computer science studies. The widespread use of actions in everyday life such as IoT, semantic web services, etc., and the limitations and issues in the action formalisms are two factors that lead us to study how actions are represented.

\textbf{Design/methodology/approach -}
Since 2007, there have been some ideas to integrate Description Logic (DL) and action formalisms for representing both static and dynamic knowledge. Meanwhile, time is an important factor in dynamic situations, and actions change states over time. In this study, on the one hand, we examined related logical structures such as extensions of description logics (DLs), temporal formalisms, and action formalisms. On the other hand, we analyzed possible tools for designing and developing the Knowledge and Action Base (KAB).

\textbf{Findings -}
For representation and reasoning about actions, we embedded actions into DLs (such as Dynamic-ALC and its extensions). We propose a terminable algorithm for action projection, planning, checking the satisfiability, consistency, realizability, and executability, and also querying from KAB. Actions in this framework were modeled with SPIN and added to state space. This framework has also been implemented as a plugin for the Protégé ontology editor.

\textbf{Originality/value -}
During the last two decades, various algorithms have been presented, but due to the high computational complexity, we face many problems in implementing dynamic ontologies. In addition, an algorithm to detect the inconsistency of actions' effects was not explicitly stated. In the proposed strategy, the interactions of actions with other parts of modeled knowledge, and a method to check consistency between the effects of actions are presented. With this framework, the ramification problem can be well handled in future works.

\end{abstract}

\end{frontmatter}

\thispagestyle{empty}
\pagestyle{empty}

\section{Introduction}
Many of the knowledge modeling and representation in our everyday lives are in dynamic knowledge structures. For example, in a web service architecture that contains a requester, a broker, and a provider, we are faced with step-by-step actions, and we can illustrate and represent the web service compositions in different case studies based on semantic webs \cite{zhang2013dynamic}.

There are many ways that dynamic-temporal knowledge can be described, modeled, reasoned about, and implemented \cite{van2008handbook}. In a related work section, we will briefly review works about action formalism in formal logics, computer science, and artificial intelligence.

Change and time formalism has a very long history from Aristotle, Diodorus, and early Islamic logicians such as Avicenna, up to modern logicians. Change formalism, mostly studies in the field of “temporal logics”. In temporal logic, using one of the quantified or modal approaches, we formalize time in either point-based or interval-based time.

The hybridization of action and change formalism, especially in dynamic knowledge base projects, is done in different methods in many of the formal logic and AI fields. Some projects, introduce dynamic logics and explicitly formalize the notion of actions and states, whereas the notion of time and change are implicit in them \cite{chang2007dynamic}. Other works, introduce temporal logic and formalize the notion of time and change, while the notion of action and dynamic behavior of the knowledge base is implicit in them. In this paper, we introduce an individual-based algorithm for description dynamic logics that implicitly contains the time notion.

Unlike ontology versioning, we don’t track the ontology changes manually, but the actions automatically cause a new version of the knowledge base to appear. Therefore, we do not only have the ontology versions. The "changes", semantically, with the help of actions within the ontology itself, cause communication between two different states.

Ontology versioning ignores the parts of an ontology that are constant and haven’t changed. In this paper, the entire new knowledge base produced by changes in actions will be examined by projection, and all realizability, executability, and planning services will be provided using TBox and ABox reasoning.

\section{Related Works}

\subsection{Action Formalism}
The logic of actions was introduced in philosophical and logical works first. The studies about the concept action started by St. Anselm in \textit{De Grammatico} were in logical context \cite{henry2012commentary}, and in two centuries, developed by Alan Ross Anderson\cite{anderson1962logic}, Frederick B. Fitch \cite{fitch1963logical}, Stig Kanger \cite{kanger1957new}, Brian F. Chellas \cite{chellas1969logical} and reported these works historically by Segerberg \cite{segerberg1992getting} and Belnap \cite{belnap2001facing}. “\textit{stit theory}” \cite{belnap2001facing} and “\textit{dynamic logic}” \cite{fischer1979propositional} are two different logical approaches to the logic of action.

Also, action formalism played a certain role in computer sciences, artificial intelligence, and linguistics. In linguistics, there are two ways in which actions play a role: On the one hand, utterances are "actions" (lead to the study of “\textit{speech acts}” \cite{austin1975things, searle1969speech}), and on the other hand, they can be used "to talk about actions" (lead to the study of the “semantics of action reports”\cite{davidson1967logical}). Also, “dynamic semantics”, where meanings are not considered as state descriptions, but as changes in the state of a hearer \cite{gardenfors1988knowledge}, is an important field of study about actions in linguistics \cite{sep-logic-action}.

Davidson \cite{davidson1967logical} gave an account of action sentences in terms of what is now widely known as "\textit{events}". The basic idea is that an action sentence has the form $(\exists e)(...)$, where \textit{e} is a variable over acts \cite{sep-logic-action}. For example, “\textit{Brutus violently stabbed Caesar}” is translated, ignoring tense, as:

\begin{equation}
(\exists e)(stab(e,Brutus,Caesar) \land violent(e))
\end{equation}

\par This allows us to capture the fact that this sentence logically entails that Brutus stabbed Caesar. This idea has been widely adopted in linguistics; Moreover, it is now assumed that basically, all verbs denote events. Thus, action sentences are those that speak about special types of events, called eventualities.

Vendler \cite{vendler1957verbs} classified verbs into four groups, and Moens and Steedman \cite{moens2005temporal} add a fifth category:

\begin{enumerate}
    \item states (“know”, “sit”)
    \item activities (“run”, “eat”)
    \item accomplishments (“write a letter”, “build a house”)
    \item achievements (“reach”, “arrive”)
    \item points (“flash”, “burst”)
\end{enumerate}

\par The first group of verbs, i.e., \textit{states}, refer to actions, and 2-5 all refer to changes.

In computer sciences, executing programs and action statements, need a logical action formalism, and this fact was realized by Turing \cite{turing1949report} and Von Neumann \cite{goldstine1947planning}. Distinguishing and studying fundamental notions such as \textit{state} by McCarthy \cite{mccarthy1993towards}, and also the development of semantics of programming languages, and program correctness, i.e. how to prove the correctness of a program, by Floyd \cite{floyd1993assigning}, Naur \cite{naur1966proof}, Hoare \cite{hoare1969axiomatic}, Dijkstra \cite{dijkstra1976discipline} and de Bakker \cite{bakker1980mathematical}, are the beginning of action formalism in computer sciences and AI. 

Representing and reasoning about actions and propositions, and description and specification of intelligent agents (in two single-agent and multi-agent approaches) are the most important fields of action formalism in AI, especially in the Semantic Web. However, methods and technologies for the handling of dynamic aspects in the semantic web, are largely missing in many of the works. To solve this problem, Käfer and Harth \cite{kafer2018rule} used Abstract State Machines (ASMs) as the formal basis for dealing with “changes” in Linked Data (which is the combination of RDF and HTTP) and introduced a rule-based programming of user agents in them. By formalization of Linked Data resources that change state over time, they present a syntax and operational semantics of a small rule-based language for these aims. They, also, presented smart application scenarios, e.g., a rule-based control for building automation, and showed their approach to be Turing complete. Käfer and Harth tried to add a formal notion of (1) an internal state in the agent, and (2) a notion of goals and capabilities to their ASMs, in their next works, such as \cite{kafer2018specifying} and \cite{herrera2019btc}.

Also, a challenge in dynamic semantic web and action formalism was the lack of efficiently generating and validating workflows that contain large amounts (hundreds to thousands) of individual computations to be executed over distributed environments. Gil et al. \cite{gil2007wings} describe a new approach to workflow creation that uses semantic representations to describe compactly complex scientific applications in a data-independent manner (e.g. in an earthquake simulation workflow). They have implemented this approach in Wings, as a workflow creation system that combines semantic representations with planning techniques. Oinn et al. \cite{oinn2007taverna} aligned a workflow system with the life sciences community in bio-informatics as a piece of knowledge with large-scale databases by $Taverna$ and the $myGrid$ to build a workflow environment to allow scientists to perform their current bio-informatics tasks in a more explicit, repeatable, and shareable manner.

\subsection{Time Formalism}
Historically, the notion of change, in correspondence with the notion of time, was first discussed by Zeno in the famous flying arrow paradox and developed by Aristotle by the notion of "\textit{future possibility}" in \textit{Organon} \cite{2013organon}. This notion is formalized often by temporal approaches, both in traditional and modern logics. There are several temporal logics defined by four major approaches: Quantified temporal logic (e.g. Avicenna, Bertrand Russell, Willard Van Orman Quine, Rescher, etc.), modal temporal logic (e.g. Aristotle, Chrysippus, Diodurus, Avicenna, Peter T. Geach, Arthur Prior, Nicholas Rescher etc.), metric and real-time temporal logics \cite{rescher2012temporal, Montanari1996metric, Montanari1996decidability, Bresolin2013metric} and hybrid temporal logics. Also, logicians modeled time in temporal logic as instant-based \cite{rescher2012temporal} or interval-based \cite{Bresolin2013metric}.

The scope of time and change formalism is very broad and we hint at it briefly, for instance, to AI and computer scientists that distinguish between "\textit{fluents}", which are propositions describing states of the world that may change over time, and "\textit{events}", representing what happens in the world and causes changes between states. Also the "\textit{frame problem}"\cite{mccarthy1981some} and the concept of "\textit{persistence}" \cite{mcdermott1982temporal}, along with formalization of them in logics were the most interesting issues that attracted computer scientists to temporal logic. In addition to instant-based and interval-based approaches to the formal model of time, there are different approaches to modeling the nature of time and change. for example, some logicians modeled time as a discrete, dense, or continuous fact. Also, we can discuss time by asking does it has a beginning or an endpoint or if is it linear, branching, or circular.

Dynamic logics, although implicitly, formalize time and change, and therefore we face an important problem about time and action formalisms: Is temporal logic expressible in dynamic logic? Or vice versa? On the one hand, Gergely and Úry \cite{gergely2012first} in the discussion about the Temporal Characterization of programs and actions, reply that as is well known, temporal logic can be defined in dynamic logic \cite{harel1984dynamic} in the propositional case. This fact was used by Meyer \cite{meyer1980ten} when he stated that temporal logic is uninteresting from a scientific point of view. They proved a theorem that says in the first-order extension, temporal logic cannot be embedded into dynamic logic \cite{gergely2012first}. Also, they proved another theorem that says in the first-order extension, dynamic logic cannot be embedded into temporal logic \cite{gergely2012first}, and therefore dynamic logic and temporal logic are incomparable.

\subsection{Time-action Formalism}
The incomparability of first-order temporal and first-order dynamic logics is an important problem in action and change formalism. By these considerations, some hybrid second-order and first-order formalisms were introduced to reasoning about action and changes, such as Sandewall \cite{sandewall1992features} and Doherty \cite{van2008handbook, doherty1994reasoning}. Some others, introduce action or change (time) formalism by propositional approach; In PDL (propositional dynamic logic) \cite{foo2002dealing}, and in propositional temporal logic \cite{calvanese2002reasoning}, for instance.

Also, does the incomparability problem hold in the description dynamic logic and temporal logic? We can see several works in description dynamic logic that explicitly formalize actions and implicitly capture temporal aspects of TBox and ABox in ontologies. Wolter Zakharyaschev \cite{wolter1998dynamic} introduced a dynamic description logic, \textit{PDLC}, that 'Actions' were used as modal operators to construct both concepts and formulas and concepts with dynamic meaning could be described, like this:

\begin{equation}
Mortal \doteq LivingBeing \sqcap <die>\neg LivingBeing
\end{equation}

\par This logic does not have an efficient decision algorithm. 

Baader et al. \cite{baader2005integrating}, introduced a dynamic description logic and studied the “\textit{projection}” and “\textit{executability}” problems. A major problem of this formalism is that actions are restricted to be either atomic actions or finite sequences of atomic actions. Shi et al. \cite{shi2005logical} introduced a logic foundation for the semantic web based on \textit{ALC}. Despite the expressive power of this logical system, this work lacks an efficient decision algorithm that is capable of the open world, and Shi tried to solve this problem in his next works. Chang et al. \cite{chang2007dynamic} introduced a dynamic description logic based on \textit{ALCO@} by developing a prefixed tableau calculus for \textit{DALCO@}. They proposed a satisfiability-checking algorithm based on this tableau by Shi et al. \cite{shi2005logical} and Chang \& Shi \cite{chang2007dynamic}. Despite the advantages of \cite{chang2007dynamic} and effectively carrying out the four reasoning tasks on actions, i.e., “\textit{realizability}”, “\textit{executability}”, “\textit{projection}”, and “\textit{Planning problems}”, we can see that this prefixed tableau calculus is not implemented in semantic web works until.

Many of the works in dynamic formalism, using the causal laws and their interplay with domain constraints, tried to solve an important problem in action formalisms, i.e. “\textit{The ramification problem}” \cite{mccain1995causal, lin1995embracing, thielscher1997ramification, denecker1998inductive, giordano2000ramification, giunchiglia2004nonmonotonic}. These works are either based on first-order and higher-order logics, like the Situation Calculus and the Fluent Calculus, and so do not admit decidable reasoning, or are decidable, but only propositional, like logics based on propositional dynamic logics or based on propositional temporal logics \cite{chang2007dynamic}.

In a decidable fragment of first-order logic, Baader et al. \cite{coelho2010verifying} used causal rules and causal relationships to resolve the ramification problem and causality, to ensure the consistency of TBox of the resulting state after the action execution, and to exploit a fixed-point semantics. Giordano et al. \cite{giordano2016asp} introduced an action formalism with $\mathcal{EL}^{\bot}$ description logic based on temporal answer sets to solve the “\textit{ramification problem}”, and to achieve this goal, they defined a temporal logic programming language that contains \textit{linear time temporal logic (LTL)}’s operators and dynamic LTL’s operators, and we can reason about complex actions and infinite computation in this formalism.

The high complexity in more implementations of state-based dynamic logic formalisms, such as \cite{baader2005integrating} and \cite{chang2007dynamic} on the one hand, and the challenges of frame problem, and insufficient and non-standard formalism in time-action approaches, and the lack of an efficient automation system and logical vulnerability, on the other hand, led some logicians and AI scientists such as \cite{10.1007/978-3-030-49161-1_31} to a 3-level formalism (containing ontological, logical and analytical levels), similar to the "Semantic Web Layer Cake" \cite{passin2004explorer}, and efficient usages of parameters as side effects, to solve this challenges.

\subsection{Dynamic Temporal Logics in Semantic Web}
Ontology management often uses ontology change management. In many applications, it is necessary to keep the ontology changes over time in order to make historical queries possible on time-varying ontologies.

\textit{Ontology Versioning}, in Semantic Web, is an important notion. A dynamic environment can be modeled with the help of ontology versioning and actually maintaining different versions of an ontology. In ontology versioning, the history of ontologies is kept only based on time, and actions do not lead to temporal changes in ontologies. So far, research and implementations have been done in the field of ontology versioning.  Some of these have compared the consequences of changes in an ontology. Another group of research has compared two different versions of the ontology. \textit{SemVersion}, which is a Protégé plugin, provides structural and semantical versioning for RDF models. In SemVersion, what causes a new version of the ontology may be manually added to it, so the actions do not structurally cause changes in the ontologies.

Marion et al.\cite{benevides2020propositional} introduced a framework called \textit{STRIPS}, which can perform reasoning on actions using a new dynamic logic derived from propositional dynamic logic, but they did not provide a planner based on the proposed dynamic logic. Unlike PDL atomic programs, which do not mention actions, actions in their framework have some pre-conditions and post-conditions.


\section{Borhan description Temporal Dynamic (DTD) Framework}
To take advantage of both the action and time formalisms in the semantic web, the Borhan DTD framework is proposed. In fact, in this framework, the actions are seen in the context of time, and each action affects the state space and moves the current state to a new state. The DTD framework forces the movement to be between the possible worlds.

In this section, to introduce the DTD framework, first, some definitions such as symbols and principles are provided, and then the temporal dynamic services are discussed.

\subsection{DTD Symbols and Principles} \label{DTD Symbols and Principles}
In DTD framework, we use the description fragment of first-order logic (i.e. Description Login). Thus, the symbols and principles of DL are used. Baader et al. \cite{baader2003description} have introduced temporal extensions in point-based and interval-based structures. Also, we use the symbols and theorems that are commonly used in temporal dynamic logic.

Considering all, we present our symbols and definitions as follows:

\begin{itemize}

\item	\textbf{State}

{A state is the knowledge set, including classes (such as the class of action, human, etc.), relations, and individuals (including instances of actions) in the structure of DL. In a state, we are in a static structure, and any member of the DL family can govern the state. As a result, based on the selected DL reasoner, the hidden data in that state can be inferred. In the structure of DL, we call it \textit{Asserted-State} when only asserted knowledge is considered, and \textit{Inferred-State}, when asserted and inferred knowledge, are considered.

Additionally, we go from one state to another with the execution of actions. We call the state before doing each action, the preliminary-state w.r.t. that action, and the state after that, the result-state w.r.t. that action.

In this framework, each state is an OWL-DL model, which may be considered \textit{ALC} as the DL governing the model, \textit{SHOIQ}, or so on. The key point is that for each state the inconsistency can be checked, and its classical logic consequences can be determined independently in the selected DL-reasoner.}

\item	\textbf{Flow of Time}

{In terms of time, to examine the effects of actions, the temporal description logic should be used. For temporal-DL, in the Handbook of description Logic \cite{baader2003description} \textit{"Flow of Time"} is defined. This concept, which we display with the symbol ${T}=\langle T,< \rangle$, is a set of time points \textit{T} that are explicitly or inferentially lead-late relationships between both of them. As a result, there is a strict linear order, and between both time points, there will be a ratio of precedence, delay, or synchrony.

In our framework, we will only deal with Discrete and Linear time structures. The Branching or Unbounded time structures are ignored in this step of the framework.

The noteworthy point is that this flow of time itself can be calculated and deduced in the DLs. For example, by stating the transitivity of being before and after, the time relationship between many time points that do not have an explicit time relationship of being before or after is deduced. Also, by some extensions of DL-reasoners such as \textit{swrla} and \textit{swrlb}, if two moments are in the form of \textit{xsd:dateTime}, it is possible to express the explicit ratio of being earlier or later between them. Therefore, by inferencing in DL, this flow of time can be calculated.}

\item	\textbf{Point-based Temporal DL Model}

{Similar to the definition given in the Handbook of Description Logic \cite{baader2003description}, we define the Point-based temporal model as follows: \textit{M=(T,I)} is a point-time model if there is a time flow $T=(T,<)$ and, for each time in \textit{T}, the function \textit{I} is $I(t)=(\Delta^{(I,t)},.^{(I,t)})$ exist.

From a conceptual view, for each point in time, we want to have a model in DL. Consequently, in the proposed framework in the semantic web language, a Point-based temporal model is a set of OWL-DL models, where for each t in T, there is an OWL-DL.
}

\item   \textbf{Action Rules}

{In our knowledge, there may exist actions that change the world when they occur. Basically, due to these changes, the modeling of classical logic alone is not sufficient and, we need appropriate action formalisms.

Generally, for any action that is going to take place and make a change in the world, the formalism of Eq.\ref{st} will be proposed and the following components can be assumed:

\begin{itemize}

\item  \textbf{Prerequisites}

{At the moment when the action happens, it may require prerequisites that if they exist, the effect will occur, otherwise, no matter how long that event lasts, there will be no effect and change. These prerequisites may be conditional on the existence of constraints or the absence of them. For example, when we turn on the light switch, the room will light up only if the building's electricity is connected and there is no curtain over it. When these conditions are met, the actions will have effects on the world and will make the necessary changes.}

\item \textbf{Relation to the Time (How that action is related to the time of effect)}

{In non-gradual and non-fuzzy actions, it is possible to determine the moment when that action makes its impact at that time in the world. For example, in instantaneous actions, their effects will take place exactly at the time of occurrence of action. Also, in actions with a time interval, changes in the state may take place at the beginning or end of the interval.

Every time an action occurs, it changes the world. So, in expressing the rule of changing, the connection of actions to the time must be on a general time variable and not a specific time.}

\item  \textbf{Effect}

{The change that takes place in the world after an action occurs is called the effect of that action. In general, both necessary effects and possible effects are considered in action formalisms. Considering the importance and applicability of necessary effects and also considering that possible effects can be modeled by negating necessity, we have considered modeling only necessary effects in our proposed framework for simplicity.

The considerable point is that we used prerequisite and effect instead of post-condition and pre-condition, because when the action occurs, the new state is created from the previous state; not that all possible states have been created, and then we want to consider changing states over them. This idea becomes the basis for solving the ramification problem, which we will mention in future works.}

\end{itemize}

\begin{equation}
\label{st}
\forall x_i\ \forall t \in T \ A(v_1, ..., v_k, x_1, ..., x_n, t) = (P, E)
\end{equation}

In our framework, Borhan DTD, the actions are ontologically placed in the OWL model, and rules in SPIN are used to define the way to effect and change. This type of rule is called a state changer. The prerequisites of each action are placed in the WHERE section of the \textit{SPARQL} command. If they are non-existential conditions, they appear in the \textit{FILTER NOT EXISTS} part. The way to relate that action to time which is actually one of the prerequisites is described in the \textit{WHERE} section as a triple. The subject of this triple, the action, its predicate is the Relation to the Time (such as \textit{"it has time"}, \textit{"it has a starting point"}, \textit{"it has an endpoint"}, etc.) and its object is \textit{?\_T} as a general variable. Conceptually, this triple says that if this action occurs at a time \textit{?\_T}, it will have these effects. Eventually, the effect of each action seems in the CONSTRUCT part of the rule.
}

\end{itemize}


\subsection{Borhan DTD Architecture}

According to the dynamic space, there are four main temporal dynamic services or reasoning tasks. Each of them is provided to respond to a specific need and can be executed independently without any specific time order. The definition of each service will be described in this section.

Unlike ontology versioning, in temporal dynamic knowledge representation, both the fixed part of the ontology and its changed part are examined in the reasoning task, due to the occurrence of actions, so all the realizability, executability, and planning services are calculated after TBox and ABox reasoning on the entire ontology.

\subsubsection{Projection}

DTD-Projection consists of two main components:
\begin{enumerate}
    \item Description Logic Reasoner Component
    \item Temporal Dynamic Logic Reasoner Component
\end{enumerate}

Each component performs the reasoning operations with the help of a corresponding logic. In other words, the Projection consists of two separate sequential logical parts in a cycle. Initially, it starts from the first state, and only at first, if the TBox reasoner finds any unsatisfiability, it will report. Because the temporal-dynamic logic rules will not change the terminology, and as a result, there will be no change in the TBox, and there is no need to re-execute the TBox reasoner.

First, a preliminary inference based on description logic is performed. Contrary to TBox reasoning, ABox reasoning will be executed in each cycle. DL-reasoning is needed because each state must be fully consistent and its logical consequences must also be included in the temporal-dynamic logic rules. Furthermore, one of the outputs of the first component is to find the flow of time (as described in section \ref{DTD Symbols and Principles}) because the second component requires knowing the sequence of time states.

The second component's task is to project temporal dynamic logic rules. In this stage, a time counter indicates where the projection process is in the flow of time. By specifying the current time and by receiving the inferred model based on the DL, the temporal-dynamic logic rules will be executed at that time. In other words, for this purpose, the \textit{?\_T} variable will be replaced with the current time, and SPIN rules will be executed. Finally, the End-Time Point Checker is responsible for checking whether the projection has reached the end of the flow of time.

\begin{figure}
\centering
\includegraphics[width=10.5 cm]{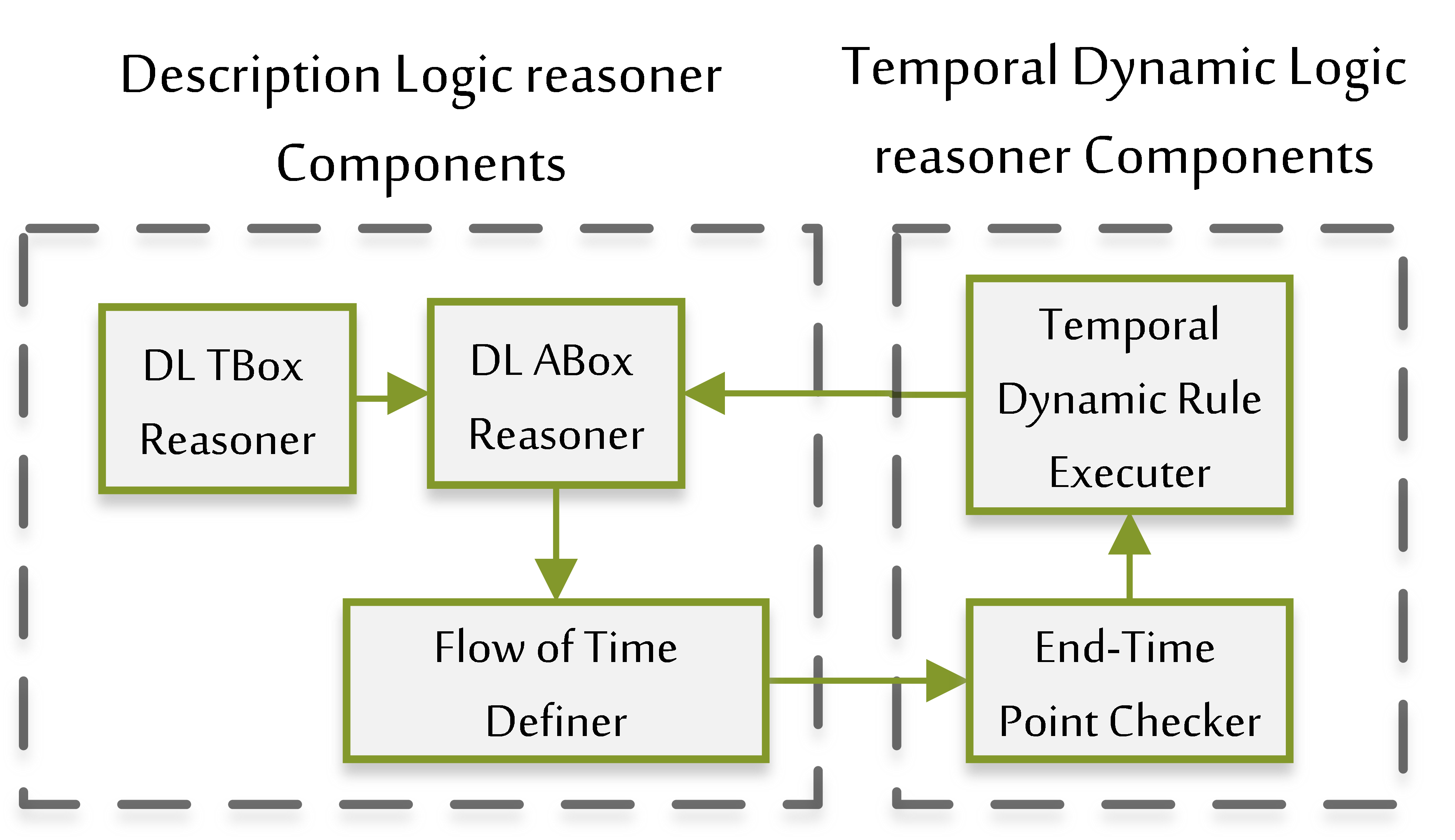}
\caption{Description and Temporal Dynamic Logic reasoner Components}  \label{arch}
\end{figure}   

\unskip\textbf{}

\subsubsection{Realizability}

According to the definition given in \cite{chang2007dynamic}, realizability is to check whether an action is reasonable or not. We define realizability in the way that no unsatisfiability has occurred in the effect of any of the rules.

\begin{figure}
\centering
\includegraphics[width=10.5 cm]{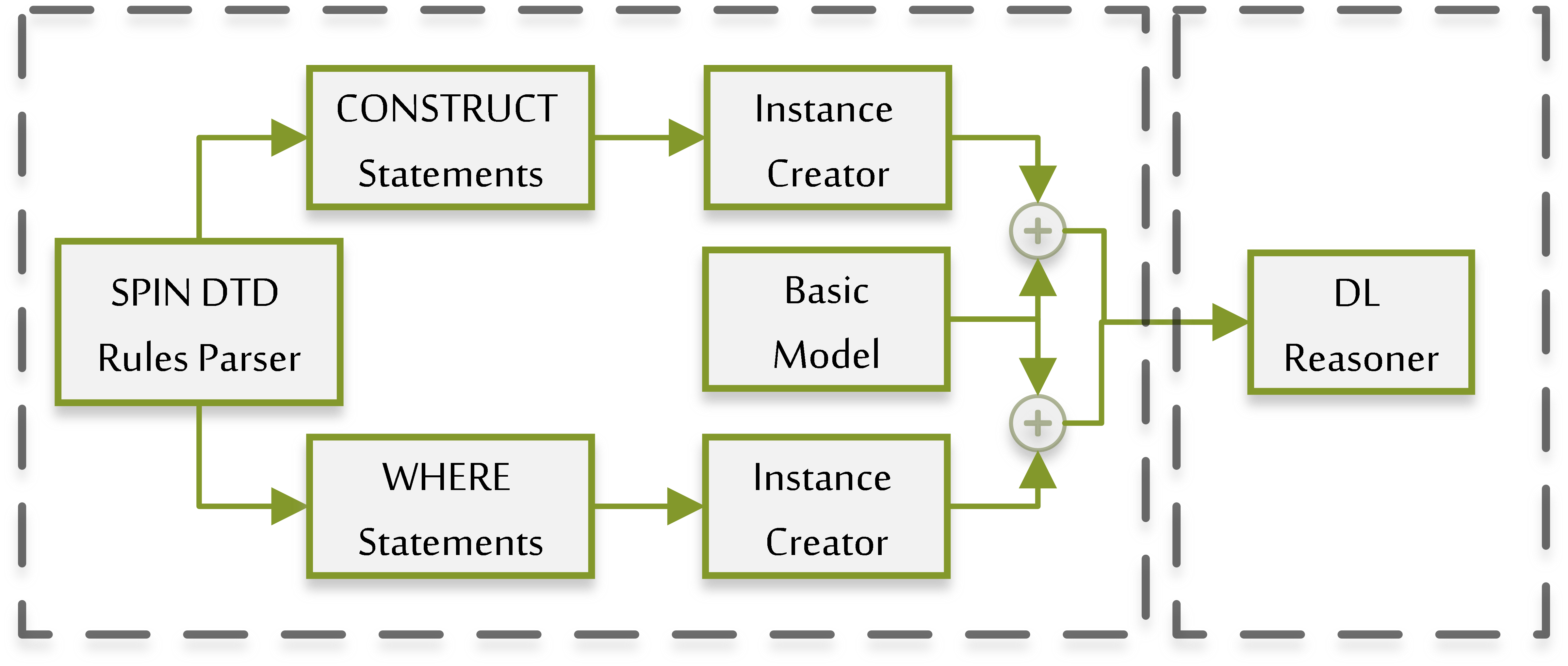}
\caption{Temporal Dynamic Logic Services - Executability and Realizability Checker} \label{real-exec-arch}
\end{figure}

Figure~\ref{real-exec-arch} shows the architecture of two realizability and executability services. At first, \textit{\textbf{SPIN DTD Rules Parser}} component receives all the defined rules, and separates them from each other into two parts: construct (i.e. effects) and where statement (i.e. prerequisites). In the realizability service, the \textit{construct} statements are given to the \textit{\textbf{Instance Creator}}, and it replaces all variables inside the triples with instances. In other words, we can say that individuals have been defined for them. The \textit{\textbf{DL Reasoner}}, having new triples, and the \textit{\textbf{Basic Model}} that contains the ontology and all class relationships, makes inferences based on description logic and reports if there is an unsatisfiability between the triples.

\unskip

\subsubsection{Executability}
All defined rules (both dynamic and non-dynamic) may consist of more than one sentence, therefore, there is a possibility of unsatisfiability between these logical sentences.

In executability, the goal is to check the executability of the actions by checking the compatibility of the prerequisites of the rules. In other words, make sure that there is no unsatisfiability in the preparation of all the rules. Therefore, we can check whether an action is applicable or not according to what is stated in the prerequisites of its rule.

As mentioned earlier, the bottom part of Figure~\ref{real-exec-arch} represents the executability service architecture. In the executability service, the \textit{where} statements are given to the instance creator, and similar to what was said, variables are replaced with instances. Here, the DL Reasoner checks the existence of unsatisfiability in the effects of a rule.

\subsubsection{Planning}

The planning technique is the process of checking the achievement of the goal, which includes a sequence of actions that are performed to reach a goal state on an initial state, and at the end provides the path/paths to reach the goal as an output. Among the applications of planning, we can mention robots, decision-making, contradiction detection, etc.

In the process of planning to provide the path to achieve the goal, the goal, and also the conditions to achieve it, may not be found directly among the rules. In this situation, TBox reasoning can be helpful. In this way, the planner has full knowledge of the definition of existing classes in ontology and their relations.

In dynamic modeling, we face multiple worlds. This means that by performing any action, we are transferred from one world to another world, where we see changes in the effects of the actions performed on the triple statements. The inference process and finding the path to the goal state are faced with many time complications in such a wide state space. In other words, the graph created in dynamic domains is much larger than in static domains, and graph navigation faces challenges. As a classic example, Sirin et al. in \cite{sirin2004htn} use the HTN (Hierarchical Task Network) planning technique to composite Services automatically. To achieve this aim, they used HTN planning system SHOP2 with OWL-S Web Service descriptions and executed the resulting plans over the Web.

In some recent research\cite{jiang2018task}, ABox reasoning-based planning algorithms were presented, which are useful for applications such as robot guidance. These reasoning methods are considered incomplete in the case of the knowledge base and ontologies containing many rules, such as what we see in this paper. Therefore, one of the most important differences of the presented dynamic planner is the ability to TBox reasoning.

In the formal definition of planning, we have the triple $(S_0, \mathcal{A}, S_g)$, which respectively shows the initial state, the set of actions defined in the preconditions, and the goal state.

The planning function receives as input the items that can be seen below along with an example:
\begin{enumerate}
    
\item {The initial state}

\item {The agent must be an individual who exists in the knowledge base.}

\item {A set of actions under the authority of the agent (example: washing, combining, etc.). These set of actions may be received by the user, or if not received, all actions under the Arbitrary Action category will be considered.}

\item {The goal triple (example: table t1 must be cleaned)\footnote{We assumed that only one triple will be received as a goal.}}
\end{enumerate}

The planning function by receiving the above inputs, recursively provides the path(s) that each of them including steps to reach the goal output. For this purpose, the following steps need to be performed:

\begin{enumerate}
\item {\textbf{Checking the goal triple}

By default, \textit{subject} is one of the individuals in the knowledge base, therefore, the search is performed according to \textit{predicate} and \textit{object}.}

\item {\textbf{ Finding a set of rules that lead to achieving the triple goal }

In other words, finding the rules that have triple targets in their effect section.}

\item {\textbf{Checking the set of prerequisites (conditions) in each of the found rules}

For each of the rules found, there is a separate path to reach the goal. The steps of these routes are recursively obtained by repeating the mentioned items. There is a counter in each path that specifies its steps recursively.}

\item {\textbf{Checking each of the conditions and choosing a set of conditions for the next round of execution}

Among each rule condition, the following items have a specific status and they are set aside for the next steps of execution.}

\begin{itemize}
    \item {A condition that contains a variable that exists in the effect part of that rule. This variable takes the individual value of the input and its status is known.}
    
    \item {The object that is the performing agent is received in the input.}
    
    \item {All variables that are members of one of the optional actions (using \textit{rdf:type} as predicate). Therefore, all triples that have such variables can be produced. Even in these triples, variables that have not yet been assigned can be generated because they can be created after their status is cleared.}
    
    \item {All variables are members of one of the classes where there is at least one individual for that class in the inferred knowledge graph. But it should be noted that triples that only one of their objects or subjects is one of these individuals is not discarded.}
    
    \item {A condition that contains the time variable \textit{?\_T}. This variable is placed based on the order of the algorithm (counter), for example, in the first step of the path, Tn, which is the last action, is placed and is reported at the end of the algorithm.}
\end{itemize}
\item {\textbf{Repeating the above steps for each of the conditions (triples) that are members of the selected set and, in other words, do not have a specific status.}

For these triples, two situations may occur:
\begin{itemize}
    \item {If a condition is found in a rule effect, the above steps are repeated, and finding the next step starts from the current path. In this way, the desired condition is assumed as a target triple and is searched among the effects of the rules.}
    \item {If the triple is not found in any rule effect, the current path is declared closed, and finding the steps of that path does not continue.}
\end{itemize}
}
\end{enumerate}


\section{ActionBox Unsatisfibity Checker}
In DL, there are appropriate tableau algorithms in TBox to check the unsatisfiability between the description rules and report it with its explanations. But in the Dynamic domain, checking unsatisfiability between the effects of actions is more complex. For example, consider that we want to model the sports rules of a school, and in the simplest case, we have to model these rules:

\begin{itemize}
    \item{When someone registers for volleyball, he joins the team.} 
    \item{When someone registers for basketball, he joins the team.}
    \item{If someone is a member of the basketball team, he will not be a member of the volleyball team, and vice versa.}
\end{itemize}

At first sight, we can't find the unsatisfiability between the action-rules. But after applying these rules to some stories, inconsistencies will appear. For example, when Ali registers in both volleyball and basketball and becomes a member of both teams, the inconsistency between the rules will arise. Having this complete story, we realized we had to rewrite the first and second rules as follows:

\begin{itemize}
\item {If someone is not a member of the basketball team and registers for volleyball, he becomes a member of the volleyball team.}
\item {If someone is not a member of the volleyball team and registers for basketball, he will become a member of the basketball team.}
\end{itemize}

So far, few attempts have been made to discover the unsatisfiability between dynamic rules without the presence of any story, but in most of these methods, we face computational and time complications. In the Borhan DTD framework, we use the planning algorithm to solve this problem. So, instead of forward calculation, we have backward analysis.

In this case, without any story, the framework explains the unsatisfiability between the effects of the actions. As part of this version of the Borhan DTD, an expert implements an initial state that is required to execute the planning algorithm, but it may be improved by using abductive methods in future works.

To realize this method, we put the main propositions in creating inconsistency such as owl:disjointWith and owl:complementOf in the desired order of planning, and check if there is a plan to bring the initial state to this desired order or not. The existence of a plan to achieve the goal shows that dynamic rules create inconsistencies in some cases. On the other hand, if no plan is found for any of the main predicates of inconsistency, it can be claimed that the modeled dynamic rules will not bring us from the initial state to the inconsistency state.

For example, in the above-mentioned sports' rules, assume there is at least one student in the initial state. In the existing modeled rules, there is only the relationship owl:disjointWith to make inconsistency. Therefore, we want that student to be a member of both the volleyball and the basketball teams. Then we check to see if there is a plan that takes us from the initial state to a state where the student is a member of both teams. After executing the planning algorithm, these two action lists are returned:

\begin{itemize}
    \item {{registration in volleyball, registration in basketball}}
    \item {{registration in basketball, registration in volleyball}}
\end{itemize}

The existence of these plans shows that although this model hasn't applied to any story, these laws lead us to inconsistent states in some cases and should be modeled more accurately.

But if we run the planning algorithm with this goal and assume modified rules, no plan will be returned to achieve that goal, and this shows that our dynamic rules will not reach inconsistency in any state.


\section{Experimental Results}
\subsection{Data}
While we cannot release the source code and the dataset used in this paper due to legal, confidentiality, and commercialization constraints, we fully recognize the value of sharing research resources to advance scientific knowledge. The code, which is proprietary to our company, is an integral component of our ongoing research efforts. The dataset, on the other hand, contains sensitive information that could be harmful to individuals or organizations if released without proper safeguards.

We understand that this may limit the ability of others to reproduce our findings or build upon our work. Nevertheless, we firmly believe that protecting sensitive information is of paramount importance and must take precedence over the potential benefits of open data sharing in this instance. However, we made every effort to provide accurate information about the data, each of the reasoning tasks, and the process of extracting results to ensure reproducibility as much as possible.

\subsection{Analysis}
A detailed report of several different scenarios that can be assumed in the dynamic DLs, from famous scenarios to designed ones, is as follows in Appendix A. In these scenarios TBox is static and ABox and Action Boxes are dynamic. These scenarios can be used as an evaluation dataset to check the capability of dynamic formalisms. In Appendix A, The "Textual Form" is a brief description of the scenario and after the title of each scenario, the most important feature of that scenario is mentioned in parentheses.

Among the dynamic temporal scenarios, for example, we report the advantages of formalizing and analyzing Scenario 3 in Appendix A. "Symbolic form", "Input data" and "Expected output data" in this scenario are shown in Table~\ref{tab:scenario}. Based on the results obtained in this paper, by applying Borhan DTD algorithm in this dynamic scenario, we can manage the action executions, projection, and planning. Also, we can check the satisfiability, consistency, realizability, and executability in the lowest complexity.

To check the computability of different formalisms according to Appendix A, and especially to test the power of problem-solving of various formalisms in realizability, executability, projection and planning (as the scenario 3) in the other 7 scenarios, we select the 6 other formalisms (Table~\ref{tab:comp}), and by the various analytical and computational methods, analyzed and compared the capacities of these 6 algorithms in dynamic description logics. The advantages of this analytical study are shown in Table~\ref{tab:comp} shortly.

One of the most important points considered in the design of Borhan-DTD is to design it in a classical structure that can easily be combined with a non-monotonic reasoner. Considering that we are currently designing the prioritized default logic reasoner (Borhan-PDL), in the next work, we will combine the Borhan-DTD with the designed non-monotonic reasoner. So, in Table~\ref{tab:comp}, we add "Borhan-DTD+PDL" to compare its ability to each other.

\begin{table}[]
\caption{Representation of the scenario of opening a bank account in the formulation of Borhan-DTD}
\label{tab:scenario}
\begin{tabular}{|l|l|l|}
\hline
\multicolumn{1}{|c|}{\textbf{Symbolic Form}} & \multicolumn{1}{c|}{\textbf{Input Data}} & \multicolumn{1}{c|}{\textbf{Expected Output Data}} \\ \hline
 & - & \textbf{Realizability: OK} \\ \cline{2-3} 
 & - & \textbf{Executability: OK} \\ \cline{2-3} 
 & \textbf{\begin{tabular}[c]{@{}l@{}} \textbf{Projection:}\\ $EligibleBank(a)$\\ $\exists holds:ProofAddress(a)$\\ $OpeningB_acc(a,t_0 )$ \end{tabular}} 
 & \textbf{\begin{tabular}[c]{@{}l@{}} $t<t_0$ \\ Initial state\\ $t>t_0$\\ Initial state +\\ $B_acc(b)\wedge$\\ $holds(a,b)\wedge$\\ $B_acc_no_credit(b)$ \end{tabular}} \\ \cline{2-3} 
\multirow{-11}{*}{\begin{tabular}[c]{@{}l@{}}$ \forall x \forall t\in T  OpeningB_acc(x,t) = (EligibleBank(x)\wedge$\\ $\exists holds:ProofAddress(x)\wedge \exists holds:Letter(x)$\\ $,\exists y B_acc(y)\wedge holds(x,y)\wedge B_acc_credit(y)$\\ \\ $\forall x \forall t\in T  OpeningB_acc(x,t) = (EligibleBank(x)\wedge$\\ $\exists holds:ProofAddress(x)\wedge$\\ $\neg \exists holds:Letter(x),\exists y B_acc(y)\wedge$\\ $holds(x,y)\wedge B_acc_no_credit(y)$\\ \\ $\forall x \forall t\in T  GetLetter(x,t) =(Human(x)$\\ $,\exists holds:Letter(x))$\\ \\ $EligibleBank(x)\supset Human(x)$
 \end{tabular}} & \textbf{\begin{tabular}[c]{@{}l@{}} \textbf{Planning:}\\ \textbf{Goal:}  $B_acc_credit(y)$\\ \textbf{Initial state:}\\ $EligibleBank(a)$\\ $\exists holds:ProofAddress$ \end{tabular}} & \textbf{\begin{tabular}[c]{@{}l@{}}$GetLetter(x,t_1)$\\ Then\\ $Opening B_acc(x,t_2)$ \end{tabular}} \\ \hline
\end{tabular}
\end{table}

\begin{table}[h]
\caption{Comparing the ability of common algorithms in Representation, Reasoning, and Planning in the Dynamic Temporal Description Logic. The "-" sign means no implementation}
\label{tab:comp}
\centering
\begin{tabular}{|l|l|l|l|l|l|l|l|l|} \hline
\backslashbox{Algorithm}{Scenario} & 1 & 2 & 3 & 4 & 5 & 6 & 7 & 8  \\ \hline
\cite{baader2005integrating} & OK- & No & OK- & No & No & No & No & No \\ \hline
\cite{chang2007dynamic} & OK- & OK- & OK- & OK- & No & OK- & No & No \\ \hline
\cite{liu2008modeling} & OK- & No & OK- & OK- & No & OK- & No & No \\ \hline
\cite{zhang2013dynamic} & OK & No & OK & OK & OK & OK & No & No \\ \hline
\cite{hariri2013description} & OK & No & OK & No & No & OK & No & No \\ \hline
\cite{bataityte2020ontological} & OK & No & OK & No & No & OK & No & No \\ \hline
Borhan DTD & OK & OK & OK & No & OK & OK & OK & No \\ \hline
Borhan DTD+PDL & OK & OK & OK & OK & OK & OK & OK & OK \\ \hline 
\end{tabular}
\end{table}

Among the various challenges in front of the above scenarios, the “frame problem” and “ramification problem” are very important issues. In the legal domain, ActionBox Unsatisfibity Checker is very needed. However, the best algorithm is an algorithm that can execute the actions, project the scenarios, plan for any goals, check the satisfiability of Tbox and ActionBox, check the consistency, realizability, and executability, and also query from KAB.

\subsection{Complexity Measurement}
We will analyze the complexity of each reasoning task separately in this section.

\subsubsection{Projection}
The Projection task has a time complexity of $O(n^2)$, where $n$ is the number of rules in the knowledge base. This is because the reasoning task involves two sequential logical parts, each of which has a time complexity of $O(n)$.

The first part performs DL reasoning, which involves checking the consistency of a set of triples. The second part performs DTD reasoning, which involves projecting the effects of actions over time. Since both DL reasoning and DTD reasoning involve checking the consistency of a set of triples, their time complexity is $O(n^2)$.

\subsubsection{Realizability}
The Realizability task has a time complexity of $O(n)$, where $n$ is the number of rules in the knowledge base. This is because the task involves checking whether an action is reasonable or not. This is done by checking whether the effect of the action is consistent with the current state of the knowledge base.

\subsubsection{Executability}
The Executability task has a time complexity of $O(n^2)$, where $n$ is the number of rules in the knowledge base. This is because the reasoning task involves checking the executability of each pair of rules, which can be done in constant time. In other words, this task is done by checking whether the prerequisites of the action are satisfied by the current state of the knowledge base. 

\subsubsection{Planning}
The Planning task has a time complexity of $O(n^d)$, where $n$ is the number of rules in the knowledge base and $d$ is the depth of the search tree. This is because the planning task involves searching for a path (the sequence of actions that can be executed to achieve a given goal) from the initial state to the goal state, which can be done in exponential time in the worst case. In other words, this algorithm must search through a tree of possible actions, and the depth of this tree is exponential in the number of actions. 

\section{Conclusions}
By applying the introduced algorithm in this paper in dynamic domains, that contain change and action formalisms in ontologies, we can make reasoning and plan over static and dynamic knowledge graphs. Certainly, in the implementation of any algorithm, there are many problems and bugs. For example, high complexity in mass knowledge bases with a lot of queries leads us to modify and improve the algorithms by new creative methods so that we can have the maximum reduction in complexity.

As said above, the usual approaches in dynamic description logics are individual-based in TBox. One of our future works is to model non-individual-based (and non-agent-based) dynamic domains, at least in T-Box. Because in the real world (actual state), there are general rules and conditions that often need to be represented as non-individual-based and at higher levels of abstraction. However, we often reduce the representation of dynamic ontologies to an interaction between an individual and its environment and state, that affects the entity's behavior or function \cite{chen2007context}.

Even if we cannot implement action formalisms, such as Chang et al. \cite{chen2007context}’s prefixed tableaux algorithm for dynamic \textit{ALCO@} DL as a non-individual based one, we can reduce the complexity of reasoning in our algorithm, as one of our important goals, by exploiting exploratory methods for pruning redundant branches.

Also, in action and change formalisms scope, especially on non-deterministic actions and causal rules, we deal with “ramification problem” \cite{giordano2013reasoning}), validity in previous states, “temporal answer sets” \cite{giordano2021reasoning} and “ramification and causality” \cite{thielscher1997ramification}. However, by a decidable approach, we will focus on dynamic description logics, since, as explained above, other works in these fields are either based on undecidable reasoning such as first- or higher-order logics, like the Situation Calculus and Fluent Calculus, or are decidable, but only in propositional approach.

However, according to the design of the states in a classical structure in the Borhan-DTD, as mentioned in the logical experiments, it is possible to apply a non-monotonic reasoner in each state, and as a result, the changes and deletions are managed by it. Considering that we are currently designing the prioritized default logic reasoner in the Borhan team, in the next work, we will combine the Borhan-DTD reasoner with it and represent and handle all the issues of the ramification problem.

By introducing and extending description logic (DLs) and increasing their application in knowledge representation especially in OWL and semantic web, many shortcomings were identified that weren’t resolvable in classical DLs. So logicians and computer scientists intended to use reasoners that formalize non-certainty and non-monotonicity (and therefore defeasibility), such as abductive reasoners and default logics.

In the recent two decades, TBox \cite{halland2014tbox} and ABox Abductive problem solvers \cite{pukancova2015abductive, pukancova2017tableau, pukancova2018abox, pukancova2020aaa, hubauer2011relaxed, homola2022hybrid, hubauer2016relaxed, petasis2013boemie} were introduced and developed. In our future works, by applying a diagnostic approach in conditions and behaviors of dynamic description reasoners, we will focus on ABox abduction problem solvers in dynamic DLs and extend plannings by creative conjectures. We are trying to find an efficient algorithm by combining the tableaux-resolution algorithms of \cite{klarman2011abox, pukancova2015abductive, pukancova2017tableau, pukancova2018abox, pukancova2020aaa}, without the need to black-Box checking and calling TA. This project is now in an implementational and experimental phase in Protégé.

Another new and efficient method to optimize algorithms in dynamic knowledge graphs is “embedding” which encodes the entities, concepts, and roles from knowledge graphs into a low-dimensional vector space \cite{wu2022efficiently}. Knowledge graphs are dynamic in the real world and complete over time by the addition or deletion of triples. In the next paper, we will present a context-aware dynamic knowledge graph by embedding method, which can not only learn embeddings from scratch but also support online embedding learning in web services, as step-by-step actions in web architecture (see \cite{zhang2013dynamic} as an example). We can build the semantic web of things through a dynamic knowledge graph (see \cite{antoniazzi2019building} as an example).


\section*{Acknowledgement}
We would like to express our sincere gratitude to \textit{Hamtaa Institute}\footnote{https://www.hamtaa.org} for their generous support of our research. Their financial assistance was instrumental in enabling us to conduct the research that led to this paper. We are grateful for their belief in our work and for their commitment to fostering scientific research.


\bibliography{bibliography.bib}

\appendix

\section{\\Dynamic Temporal Description logic scenarios}

\begin{longtable}
{|>{\fontsize{8}{10}\selectfont}p{.04\linewidth}|>{\fontsize{8}{10}\selectfont}p{.12\linewidth}|>{\fontsize{8}{10}\selectfont}p{.59\linewidth}|>{\fontsize{8}{10}\selectfont}p{.1\linewidth}|}

  \caption{Famous scenarios in the domain of Dynamic Temporal Description logic and their specifications} \\ 
  \hline
  \rowcolor{Gainsboro!60} 
  \textbf{No.}  & \multicolumn{1}{c|}{\textbf{Title}} & \multicolumn{1}{c|}{\textbf{Textual Form}} & \multicolumn{1}{c|}{\textbf{Source}}
  \endhead
  \hline
1 & Blocks’ world (Projection, Planning) & This famous scenario is a planning domain in artificial intelligence. The initial situation is a set of wooden blocks of various shapes and colors placed on a table. Also, this scenario contains some restrictions as “Only one block may be moved at a time: it may either be placed on the table or placed atop another block.”. Then, any blocks that are, at a given time, under another block cannot be moved. Therefore, the next situation is generated by moving a block (moving a block changes its position, but not its color, shape, etc.). The final goal in this scenario is to build one or more vertical stacks of blocks. & \cite{hayes1974some, slaney2001blocks} \\ 
\hline
  2 & Yale shooting problem (Ramification Problem, Time handling, Projection) & \textbf{Two fluent:} being alive for turkey and being loaded for a gun. (Fluent: a condition that can change truth value over time)
\textbf{Time points:} 0, 1, 2, 3, 4.

\textbf{Initial time(0):} The first condition is true (alive(turkey,0)).

\textbf{Time Point (1):} The second is false (not loaded(gun,1)). 

\textbf{Time Point (2):} The gun is loaded (loaded(gun,2)). 

\textbf{Time Point (3):} Some time passes, and the gun is fired and the turkey isn't no longer alive. (shooting(gun,3)).

\textbf{Time Point (4):} The first condition is false (alive(turkey,4)).

By considering different sequences of actions in this example, different scenarios can be designed that in some of which the turkey survives.

What are the scenarios for the turkey to survive?

& \cite{hanks1987nonmonotonic} \\ 
\hline
3 & Opening a bank account (Projection) & Consider the actions of opening a bank account and applying for child benefits in the UK. Suppose the pre-condition of opening a bank account b is that the customer a is eligible for a bank account in the UK and holds a proof of address. Moreover, suppose that, if a letter from the employer is available, then the bank account comes with a credit card, otherwise not.

 & \cite{baader2005integrating} \\ 
\hline
4 & Computer server purchase service (Executability, Planning) & Imagine a computer server purchase service (CP) that can provide users with personalized configurable computers. CP begins with accepting a user’s order, which specifies the user’s requirements for the configuration. After the user makes a deposit, two services are invoked in parallel, namely, the Order Monitor (OM) service, and Order Host (OH) service, which are used to order the required monitor and host, respectively. After all the components are available, a Computer Delivery (CD) service is invoked. In the sequel, the user should complete the payment before the whole process terminates. 

This process may involve multiparty interests and requirements, as follows:

User requirements:

\begin{enumerate}
\item {The user wants to have at least 24 hours to complete his payment after receiving the ordered product.}
\end{enumerate}

Composite service requirements:
\begin{enumerate}
\item {CP wants the user to make payment a priori, while the bank service keeps it as a deposit.}
\item {CP wants the user to complete the entire payment within 12 hours after the ordered product is delivered.}
\item {If the user later cancels the order, CP wants to have a certain percentage of the deposit kept by the bank as compensation.}
\end{enumerate}
Computer Delivery (CD) service requirements:
\begin{enumerate}
\item{CD service promises that the computer delivery process will not exceed 12 hours after the service invocation.}
\end{enumerate}
 & \cite{liu2008modeling} \\
\hline
5 & Semantic web service composition (Optimized Planning) & Suppose there is a tourist who wants to go to Hainan from Beijing. He wants to subscribe to the service from the Internet. In this case, the traveler inputs the traveling date, and the system gives the vehicle plus the hotel booking information. The traveler wants the result according to his preference, such as the least money or least time. In this case, there are three services: booking a train ticket, booking an air ticket, and booking a hotel. However these services do not store in the same location, so they need to composite the services. & \cite{zhang2013dynamic} \\ 
\hline
6 & A super-heroes comics world (Projection, Classic reasoning) & In this scenario, we have a knowledge and action base:$K = (T, A_0, \Gamma, \Pi)$ . TBox (T) rendering a UML Class Diagram. this scenario contains a dynamic domain, a set of superheroes that fight each other like Batman and super-villains like Joker, and a set of cities. Each character lives in one city at a time and superheroes help the endeavors of law enforcement fighting villains threatening the city they live in. When a villain reveals himself for perpetrating his nefarious purposes against the city’s peace, he consequently becomes a declared enemy of all superheroes living in that city.

In ABox, the assertion form alterEgo(s,p) aims to a secret identity (A common trait of superheroes). Villains always try to unmask superheroes, i.e., find their secret identity, to exploit such knowledge to defeat them. & \cite{hariri2013description}\\ 
\hline
7 & Sports’ rules of a school (ActionBox Unsatisfibity Checker without any individuals) & Consider that we want to model the sports rules of a school, and in the simplest case, we have to model these rules:
\begin{itemize}
\item{When someone registers for volleyball, he joins the team.}
\item{When someone registers for basketball, he joins the team.}
\item{If someone is a member of the basketball team, he will not be a member of the volleyball team, and vice versa.}
\end{itemize}
 & Designed by Borhan-DTD team\\ 
\hline
8 & Mechanic (Ramification Problem, Prioritized Projection) & \textbf{Static TBox:}
\begin{itemize}
\item{Machine has a type from disjoint types (e.g., lathe, milling machine, drill press)}
\item{Machine has a maximum capacity (e.g., maximum weight it can handle)}
\item{Machine has a set of prioritized available tools (e.g., drill bits, cutting blades). Priority means that while a task with a higher priority can be executed, a task with a lower priority is not allowed to be executed.}
\end{itemize}
\textbf{Dynamic ABox:}
\begin{itemize}
\item {Machine is currently in one of three states: idle, running, or maintenance}
\item{Machine has a current load (e.g., the weight of the material being worked on)}
\item{Machine has a set of currently installed tools}
\end{itemize}
\textbf{Action Box:}
\begin{itemize}
\item{Operator can start or stop the machine}
\item{Operator can change the current load on the machine}
\item{Operator can install or remove tools from the machine}
\end{itemize}
\textbf{Scenario:}
A lathe machine is currently idle and has a maximum capacity of 500 kg. It is equipped with a cutting blade and a drill bit. (The priority of the cutting blade is higher than the priority of the drill bit).

The operator loads a piece of metal onto the machine weighing 300 kg and starts the machine. The machine enters the running state and begins to turn the metal piece using the cutting blade. After some time, the operator decides to switch to using the drill bit to make some holes in the metal piece. They stop the machine, remove the cutting blade, install the drill bit, and restart the machine with no change in load. The machine continues to run in its new configuration until it finishes working on the metal piece and returns to its idle state.
 & Designed by Borhan-DTD team\\
\hline
\end{longtable} 

\end{document}